\documentclass[letterpaper]{article} 
\usepackage{aaai23-r2hcai}  
\usepackage{times}  
\usepackage{helvet}  
\usepackage{courier}  
\usepackage[hyphens]{url}  
\usepackage{graphicx} 
\usepackage{amsmath} 

\usepackage{xcolor} 
\usepackage{url}

\usepackage{natbib}  
\usepackage{caption} 
\usepackage{algorithm}
\usepackage{algorithmic}
\usepackage[english]{babel}
\usepackage{fancyhdr}
\usepackage{newfloat}
\usepackage{listings}
\DeclareCaptionStyle{ruled}{labelfont=normalfont,labelsep=colon,strut=off} 
\floatstyle{ruled}
\newfloat{listing}{tb}{lst}{}
\floatname{listing}{Listing}
\usepackage{fancyhdr}
\fancypagestyle{firstpagehf}
{
   \fancyhf{}

   \fancyfoot[C]{\thepage}
}

\title{Co-NAML-LSTUR: A Combined Model with Attentive Multi-View Learning and Long- and Short-term User Representations for News Recommendation}
\author{
    Minh Hoang Nguyen$^{1,3}$* \quad
    Thuat Thien Nguyen$^{2,3}$ \quad
    Minh Nhat Ta$^{2,3}$ \quad \\
    Tung Le$^{1,3}$ \quad
    Huy Tien Nguyen $^{1,3}$
}
\affiliations {
    \textsuperscript{\rm 1} Faculty of Information Technology, University of Science, Ho Chi Minh City, Vietnam \\
    \textsuperscript{\rm 2} University of Information Technology, Ho Chi Minh City, Vietnam \\
    \textsuperscript{\rm 3} Vietnam National University, Ho Chi Minh City, Vietnam \\
    24C15049@student.hcmus.edu.vn, \{20521998, 20521614\}@gm.uit.edu.vn,
    \{lttung, ntienhuy\}@fit.hcmus.edu.vn
}

\usepackage{bibentry}

\begin{document}
\thispagestyle{firstpagehf}
\maketitle

\begin{abstract}
News recommendation systems play a critical role in alleviating information overload by delivering personalized content. A key challenge lies in jointly modeling multi-view representations of news articles and capturing the dynamic, dual-scale nature of user interests-encompassing both short- and long-term preferences. Prior methods often rely on single-view features or insufficiently model user behavior across time. In this work, we introduce \textbf{Co-NAML-LSTUR}, a hybrid news recommendation framework that integrates NAML for attentive multi-view news encoding and LSTUR for hierarchical user modeling, designed for training on limited data resources. Our approach leverages BERT-based embeddings to enhance semantic representation. We evaluate Co-NAML-LSTUR on two widely used benchmarks, MIND-small and MIND-large. Results show that our model significantly outperforms strong baselines, achieving improvements over NRMS by 1.55\% in AUC and 1.15\% in MRR, and over NAML by 2.45\% in AUC and 1.71\% in MRR. These findings highlight the effectiveness of our efficiency-focused hybrid model, which combines multi-view news modeling with dual-scale user representations for practical, resource-limited resources rather than a claim to absolute state-of-the-art (SOTA). Code is publicly available. The implementation of our model is publicly available at \textit{\underline{\url{https://github.com/MinhNguyenDS/Co-NAML-LSTUR}}}.
\end{abstract}

\section{I. Introduction}

Online news platforms such as Google News\footnote{\url{https://news.google.com}} and Microsoft News\footnote{\url{https://microsoftnews.msn.com}} have become widely adopted channels for accessing the latest news updates \citep{das2007google}. However, the vast amount of news generated every day makes it challenging for users to quickly find articles that match their interests. To address information overload and enhance user experience, personalized news recommendation systems have become an essential component of these platforms \cite{liu2010personalized, phelan2011terms}.

\begin{figure}[ht]
    \centering
    \includegraphics[width=0.9\columnwidth]{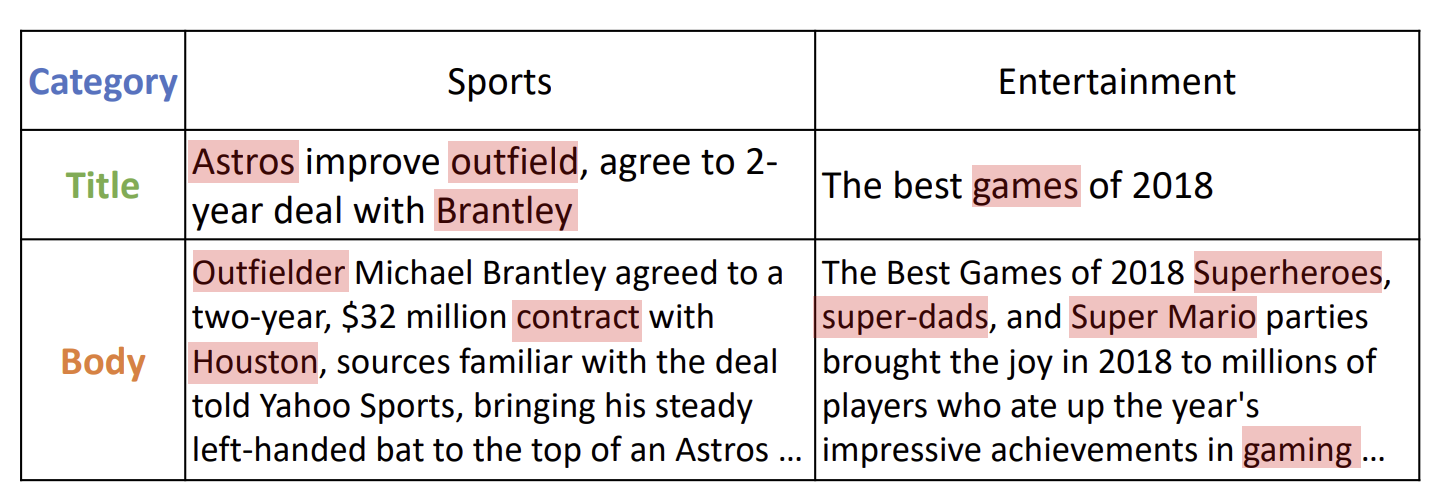} 
    \caption{Examples of two news articles that a user has read. \textit{Image adapted from studies} \cite{bib1}.}
    \label{fig_if1}
\end{figure}

\begin{figure}[ht]
    \centering
    \includegraphics[width=0.9\columnwidth]{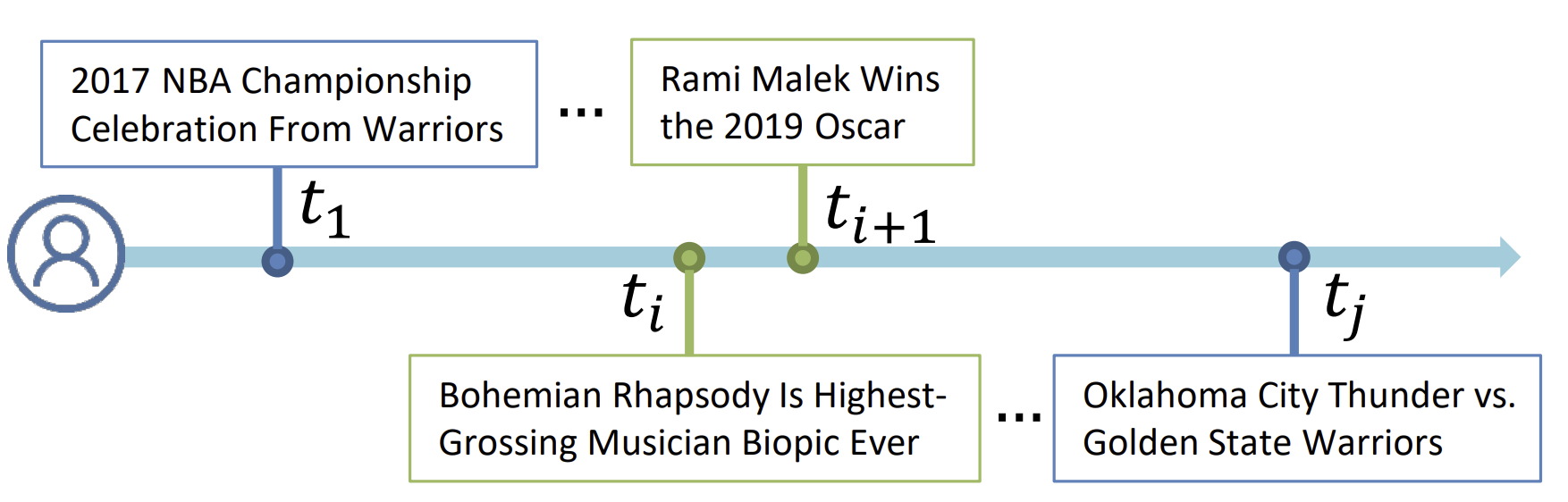} 
    \caption{An illustration of long-term and short-term user interests in news reading. \textit{Image adapted from studies} \cite{an2019neural}.}
    \label{fig_if2}
\end{figure}

The headline of the first news article in Figure \ref{fig_if1} is precise and informative, serving as a good representation of the article’s content, while the headline of the second article is shorter, more ambiguous, and less informative. Moreover, different words within the same article carry varying levels of importance; words highlighted in red denote keywords that are crucial for understanding the news. As shown in Figure \ref{fig_if2}, if a user is a fan of the “Golden State Warriors,” they are likely to read basketball-related news about this NBA team for several years. We refer to this type of preference as a \textit{long-term interest}. In contrast, user interests may also evolve over time due to specific contexts or temporary needs. For example, after reading news about the movie “Bohemian Rhapsody,” the user may begin to read related news such as “Rami Malek Wins the 2019 Oscar” because “Rami Malek” is a key actor in the film, even if the user had never shown interest in him before. We define this type of preference as a \textit{short-term interest}. Therefore, both long-term and short-term interests are critical for personalized news recommendation, and distinguishing between them helps to build more accurate user profiles.

Accurately aligning user preferences with candidate news articles is crucial for delivering high-quality recommendations. Existing methods often attempt to learn a single user representation by aggregating previously read news through deep learning models, and then match this user vector with candidate news vectors to predict relevance. For instance, \cite{okura2017embedding} proposed a method using neural networks, CNN, and GloVe to learn user and news representations by exploiting various news attributes. Other studies, such as \cite{an2019neural}, introduced GRU-based approaches to jointly model long- and short-term user interests, while \cite{wu2019npa} utilized knowledge graph representations and KCNN for news feature extraction combined with attention mechanisms to capture user preferences. However, these models typically focus on only one aspect of the recommendation pipeline and fail to effectively address other sub-tasks.

Motivated by limitations, we propose \textbf{Co-NAML-LSTUR} \textit{(a \textbf{Co}mbined model with (\textit{\textbf{N}eural}) \textbf{A}ttentive \textbf{M}ulti-view \textbf{L}earning and \textbf{L}ong- and \textbf{S}hort-\textbf{t}erm \textbf{U}ser \textbf{R}epresentations)}, a novel framework that combines the strengths of two state-of-the-art models. Co-NAML-LSTUR is designed to simultaneously learn multi-view news representations (titles, categories, abstracts), while also modeling both short-term behaviors (recently read articles) and long-term user preferences (historical clicks over time). Furthermore, we use a neural network-based scoring function for candidate news matching and leverage BERT-base word embeddings to enhance the quality of text representations. The main contributions can be summarized as follows:
\begin{itemize}
    \item We propose \textbf{Co-NAML-LSTUR}, a hybrid model that integrates attentive multi-view news representation with long- and short-term user modeling for personalized news recommendation, designed for training with fewer resources compared to state-of-the-art methods.
    \item We introduce \textbf{MIND-tiny}, a compact yet subset of MIND dataset designed to enable efficient model training on resource-constrained devices.
    \item Experimental results on the MIND show that Co-NAML-LSTUR achieves competitive performance compared to strong baselines, confirming the effectiveness of our dual-view encoder and attention-based user modeling.
\end{itemize}

\section{II. Related Work}

\subsection{2.1 News Recommendation}
The goal of news recommendation systems is to identify news articles that align with user interests from a set of candidate items \cite{das2007google}. Two key challenges are: (i) learning effective representations of news content, and (ii) modeling user interests based on their historical interactions \cite{okura2017embedding}. News recommendation is closely related to natural language processing (NLP) since news articles are textual in nature, and NLP models such as CNNs and Transformers are often leveraged to extract semantic features from news text. Furthermore, modeling user preferences from their browsing history is analogous to learning contextual dependencies from sequential text data, which has drawn increasing attention from the NLP community \cite{an2019neural}.

\subsection{2.2 Traditional Recommendation Methods}
Traditional news recommendation methods rely heavily on feature engineering to represent both news and users. In typical recommender systems, most features (e.g., \textit{userID}, \textit{itemID}) are categorical and represented as discrete IDs, leading to research focusing on modeling feature interactions. For example, Factorization Machines (FM) \cite{rw4} capture feature interactions via pairwise inner products, while Wide\&Deep \cite{cheng2016wide} and DeepFM \cite{rw5} integrate shallow and deep components for enhanced modeling capacity. Deep \& Cross Networks (DCN) \cite{wang2017deep} further model cross-feature interactions using cross layers.

However, these ID-based methods suffer from several limitations. They require extensive domain knowledge for crafting features and cannot effectively capture the rich semantic content or contextual relationships present in news articles. While most traditional methods represent a user with a single embedding vector, \cite{rw7} introduced capsule routing to extract multiple interest vectors for users. Nevertheless, this approach is specifically designed for e-commerce recommendations rather than dynamic news environments. Our objective is to model how users consume news over time in order to infer their preferences.

\subsection{2.3 Neural News Recommendation}
Traditional ID-based methods often struggle with the cold-start problem in news recommendation due to the rapid update and short lifespan of news articles \cite{rw8}. To overcome these issues, neural network-based methods have been proposed to automatically learn representations of news and users \cite{okura2017embedding}. Sequential recommendation approaches have also gained attention, aiming to model user behaviors with deep learning-based news recommendation methods, such as RNNs, CNNs, or self-attention networks \cite{wu2019npa,nguyen2024rrs,khattar2018weave,kumar2017word}. For instance, \cite{okura2017embedding} proposed to learn news embeddings from article bodies using denoising autoencoders and capture user interests using GRU networks. \cite{bib3} introduced a knowledge-aware CNN that incorporates information from knowledge graphs to encode news titles.

These approaches often rely on a single type of news information (e.g., title or body), which may be insufficient for learning comprehensive news and user representations. Our approach, in contrast, employs the multi-view news encoder of NAML to integrate diverse information sources such as titles, abstracts, and categories, providing a richer and more accurate representation of news content.

Attention mechanisms have been widely applied in neural news recommendation, including personalized attention networks \cite{wu2019npa} and multi-head self-attention \cite{rw13}, to highlight informative words or articles. More recently, pre-trained language models such as BERT \cite{rw14} have been adopted to further enhance news representations due to their superior language understanding capabilities. However, most existing methods learn a single user embedding and fail to distinguish between a user’s long-term preferences and short-term interests. In our model, we address this by employing the user encoder LSTUR,  which explicitly models both long-term and short-term user representations over time.

\begin{figure*}[!ht]
\centering
\includegraphics[width=7in]{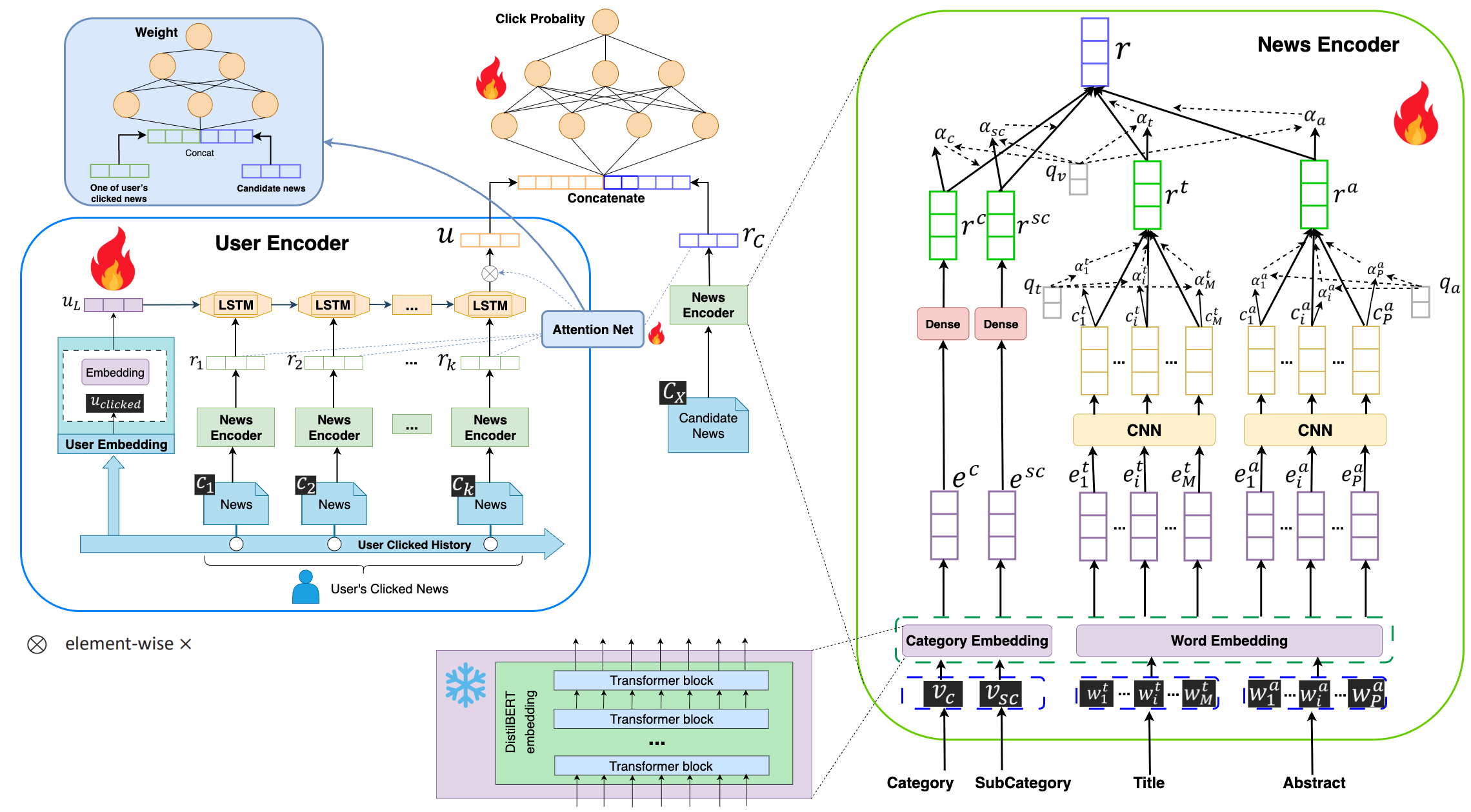} 
\caption{The framework of our Co-NAML-LSTUR approach for News Recommendation, which consists of a multi-view news encoder, a long- and short-term user encoder, and a click predictor.}
\label{fig2}
\end{figure*}

\section{III. Our Method}

In this section, we present Co-NAML-LSTUR, our proposed architecture for personalized news recommendation. The model consists of three main modules: (1) a multi-view news encoder adapted from the NAML framework, (2) a user encoder leveraging the LSTUR design to capture user interests with long- and short-term components, and (3) a click predictor inspired by the DKN model to estimate the relevance of candidate news. The overall architecture is illustrated in Figure~\ref{fig2}.

\subsection{3.1 News Encoder}

Learning high-quality news representations from multiple heterogeneous textual views is crucial for effective personalized recommendation. Inspired by the multi-view attentive structure proposed in NAML \cite{bib1}, we adopt and adapt this design for our news encoder module due to its ability to capture complementary semantics from different aspects of a news article. Specifically, we incorporate four distinct types of information: title, abstract, category, and subcategory, each processed independently to preserve structural and semantic nuances, followed by attentive fusion to produce a unified representation.

\vspace{1mm}
\noindent\textbf{Title Encoder.}
We encode news titles using a three-step architecture that includes contextual embedding, convolutional encoding, and attention-based pooling. Given a title sequence $\{w_1^t, w_2^t, \dots, w_M^t\}$, each token is first transformed into a semantic vector using a pretrained BERT-base embedding layer. These contextualized embeddings provide rich linguistic representations that serve as effective input features for downstream processing. In our implementation, we adopt DistilBERT  \cite{sanh2019distilbert} as the underlying encoder. This model is a distilled version of BERT that retains most of its representational power while being significantly smaller, faster, and more efficient—qualities particularly important for training under limited computational resources. BERT has been successfully applied in various text ranking problems  \cite{khattab2020colbert, karpukhin2020dense}, making it a practical and effective choice for encoding news titles within our recommendation framework.

Next, a 1D CNN \cite{kim2014convolutional} layer is applied to capture local semantic patterns. ReLU \cite{glorot2011deep} is the non-linear activation function. For the $i$-th word in the title, the contextual representation $c_i^t$ is computed as:

\begin{equation}
    c_i^t = \mathrm{ReLU}(F_t \cdot e_{(i-k):(i+k)}^t + b_t)
    \label{eq:cnn-title}
\end{equation}

\noindent where $e_{(i-k):(i+k)}^t$ denotes the concatenated embeddings within a fixed-size window, and $F_t$, $b_t$ are CNN kernel weights and biases respectively.

To identify the most informative words, we apply a word-level attention mechanism \cite{wu2019neural}. The attention weight $\alpha_i^t$ for word $i$ is computed as:

\begin{equation}
    a_i^t = q_t^\top \tanh(V_t c_i^t + v_t), \quad 
    \alpha_i^t = \frac{\exp(a_i^t)}{\sum_{j=1}^{M} \exp(a_j^t)}
    \label{eq:attn-title}
\end{equation}

\noindent where $q_t$, $V_t$, and $v_t$ are trainable parameters. The final title representation $r^t$ is the weighted sum of contextual vectors:

\begin{equation}
    r^t = \sum_{j=1}^{M} \alpha_j^t c_j^t
\end{equation}

\vspace{1mm}
\noindent\textbf{Abstract Encoder.}
To complement the title with a more detailed textual summary, we include an abstract encoder with an identical structure to the title encoder. Given an abstract word sequence $\{w_1^{abs}, w_2^{abs}, \dots, w_P^{abs}\}$, we use shared DistilBERT embeddings, followed by CNN and word-level attention:

\begin{equation}
    a_i^{abs} = q_{abs}^\top \tanh(V_{abs} c_i^{abs} + v_{abs}), \quad \\
    \alpha_i^{abs} = \frac{\exp(a_i^{abs})}{\sum_{j=1}^{P} \exp(a_j^{abs})}
\end{equation}

\begin{equation}
    r^{abs} = \sum_{j=1}^{P} \alpha_j^{abs} c_j^{abs}
\end{equation}

\vspace{1mm}
\noindent\textbf{Category Encoder.}
Many news platforms tag articles with hierarchical topic labels (e.g., categories and subcategories), which provide high-level semantic cues about the content. To incorporate this metadata, we encode both category ID $v_c$ and subcategory ID $v_{sc}$ via dense embeddings, initialized using DistilBERT and nonlinear projections:

\begin{equation}
    r^c = \mathrm{ReLU}(V_c \cdot \mathrm{Emb}(v_c) + b_c),
 \end{equation}

 \begin{equation}
    r^{sc} = \mathrm{ReLU}(V_{sc} \cdot \mathrm{Emb}(v_{sc}) + b_{sc})
 \end{equation}

\noindent where $\mathrm{Emb}(\cdot)$ denotes embedding lookup, and $V_c, V_{sc}, b_c, b_{sc}$ are learnable parameters.

\vspace{1mm}
\noindent\textbf{View-level Attention.}
Different views (e.g., title vs. abstract) may carry varying levels of informativeness depending on the news content. To adaptively integrate them, we use a view-level attention mechanism that assigns a relevance weight to each view. For instance, the attention score for the title view is computed as:

\begin{equation}
    a_t = q_v^\top \tanh(U_v r^t + u_v), \quad
    \alpha_t = \frac{\exp(a_t)}{\sum_{v \in \{t,abs,c,sc\}} \exp(a_v)}
\end{equation}

\noindent with similar computations for $\alpha_{abs}, \alpha_c, \alpha_{sc}$. The final unified news representation is the weighted sum across views:

\begin{equation}
    r = \alpha_t r^t + \alpha_{abs} r^{abs} + \alpha_c r^c + \alpha_{sc} r^{sc}
    \label{eq:final-news-repr}
\end{equation}

\vspace{1mm}
\noindent\textbf{Discussion.}
We inherit the multi-view design from NAML due to its interpretability and proven effectiveness in capturing complementary signals from different textual and categorical components of news articles. Our extensions include (1) replacing shallow word embeddings with BERT-base initialization to better encode contextual semantics, and (2) substituting the original "body" with "abstract" to reduce redundancy and avoid potential ethical or sensitive content. These changes allow us to retain the core strengths of NAML while improving efficiency, robustness, and suitability for our target domain.

\subsection{3.2 User Encoder}

The user encoder aims to learn comprehensive user representations by modeling both their long-term preferences and short-term interests based on historical news consumption behavior. Inspired by the LSTUR framework \cite{an2019neural}, we adopt a dual-component structure consisting of a short-term user interest module and a long-term user preference module. However, we introduce key modifications tailored to our research goals and resource constraints.

\vspace{1mm}
\noindent\textbf{Short-Term User Representation.}
Users often exhibit temporal and context-sensitive interests when reading news. To capture such short-term dynamics, we represent a user’s recent reading history as a time-ordered sequence $C = \{c_1, c_2, \dots, c_k\}$, where each $c_i$ is an article clicked by the user. Each news article is encoded into a dense vector ${r}_i$ using the News Encoder described previously.

While the original LSTUR utilizes GRU \cite{cho2014learning} for modeling short-term behavior, we hypothesize that LSTM can capture richer temporal dependencies—particularly when modeling longer user histories. Therefore, we replace GRU with LSTM \cite{hochreiter1997lstm} in our implementation. The final short-term user vector $u$ is taken as the last hidden state of the LSTM after processing the full click sequence:
\begin{equation}
    {u}_s = \text{LSTM}(r_1, {r}_2, \dots, {r}_k) 
\end{equation}

This modification is designed to enhance the model’s capacity to track evolving user preferences over time.

\vspace{1mm}
\noindent\textbf{Long-Term User Representation.}
Beyond transient interests, users also maintain stable preferences over longer periods. To model such long-term traits, we follow LSTUR’s strategy of learning user embeddings via user IDs. Specifically, for a given user ID $u$, we retrieve the corresponding long-term embedding from a trainable lookup table:
\begin{equation}
{u}_l = {W}_u[u]
\end{equation}
These embeddings are updated during training and allow the model to learn consistent behavioral traits across sessions.

\vspace{1mm}
\noindent\textbf{Attention-based User Interest Extraction.}
In addition to modeling long-term and short-term user representations, we incorporate an attention mechanism inspired by \cite{bib3} to enhance the expressiveness of user embeddings. Empirical evaluation in their research demonstrates that incorporating attention improves model performance \cite{bib3}. In the news domain, user preferences are often diverse and context-dependent, spanning a wide range of topics such as politics, technology, or entertainment. 

Therefore, assuming equal importance across all previously clicked articles may fail to accurately capture users’ dynamic interests when evaluating a new candidate article. So, we employ an attention network to assign adaptive weights to each clicked news item based on its relevance to the candidate news. Specifically, let $\{r_{1}, r_{2}, ..., r_{k}\}$ denote the encoded embeddings of a user's previously read news articles, and $r_{c}$ be the embedding of a candidate news item. For each historical article $r_{i}$, we compute an attention score $r_{i}$ reflecting its contextual relevance to $r_{c}$ as follows:

\begin{equation}
s_i = \frac{\exp(H([r_i; r_c]))}{\sum_{j=1}^{k} \exp(H([r_j; r_c]))}
\label{eq:attention}
\end{equation}

where $H(\cdot)$ is a feedforward neural network (DNN), and $[\cdot; \cdot]$ denotes vector concatenation. The final attention-based user embedding $u_{att}$ is computed as the weighted sum of historical embeddings:

\begin{equation}
u_{att} = \sum_{i=1}^{k} s_i \cdot r_i
\label{eq:att_user_rep}
\end{equation}

This attention mechanism allows the model to selectively focus on user behaviors that are more indicative of interest in the current candidate news. 

\vspace{1mm}
\noindent\textbf{Integration of Long- and Short-Term Representations.}
LSTUR proposes two methods to combine short-term and long-term representations: (1) concatenation (LSTUR-con), and (2) using the long-term vector to initialize the GRU hidden state (LSTUR-ini). Empirical results in the original study show that LSTUR-ini consistently outperforms LSTUR-con across benchmarks. Motivated by this, we adopt the \textbf{LSTUR-ini} strategy.

Specifically, we initialize the LSTM’s hidden state with the long-term embedding ${u}_l$ and update it with the recent click sequence, leading to a final user vector that reflects both persistent interests and recent behavioral trends:
\begin{equation}
    u = \text{LSTM}({r}_1, \dots, {r}_k; {h}_0 = {u}_l) \cdot {u}_{att}
\end{equation}

This hybrid design allows our model to efficiently leverage both historical and contextual cues, while benefiting from the representational power of LSTM in learning from extended user behavior. $u_{att}$ is described in Section "Attention-based User Interest Extraction".

\subsection{3.3 Click Predictor}

To estimate the probability that a user will click on a candidate news item, we implement and evaluate two representative approaches for click prediction, drawing inspiration from the NAML \cite{an2019neural} and DKN \cite{bib3} models.

\subsubsection{Dot Product-Based Predictor.} Following the approach proposed in NAML and earlier work by \cite{okura2017embedding}, we use the dot product between the user representation $u$ and the candidate news representation $r_c$ to compute the click probability:
\begin{equation}
\widehat{y} = u^\top r_c
\end{equation}

\subsubsection{Neural Network-Based Predictor.} Inspired by the DKN model, we also explore a more expressive click predictor using a feedforward neural network (DNN). This method considers richer interactions between user and candidate news embeddings. Specifically, we concatenate $u$ and $r_c$ and feed the result into a DNN $G(\cdot)$ to predict the click probability:
\begin{equation}
\widehat{y} = G([u; r_c])
\end{equation}

This formulation allows the model to capture non-linear relationships and complex interaction patterns between user preferences and candidate content. While this approach is more computationally expensive than the dot product, it has the potential to improve predictive performance, particularly when user interests are diverse and not linearly aligned with candidate representations.\\
We implement both click predictor variants in our framework to analyze their trade-offs and effectiveness. A detailed comparison of their empirical results is presented in Section 5.5 (\textit{Effectiveness of Click Prediction}).

\subsection{3.4 Model Training}

Following prior work \cite{bib1}, we adopt the Noise Contrastive Estimation (NCE) loss for model training. For each news article that has been clicked by a user (treated as a positive sample, ${{\widehat{y}}_i^+}$), we randomly sample $K$ news articles that were not clicked by the same user (treated as negative samples, $[{\widehat{y}}_1^-, {\widehat{y}}_2^-, \dots, {\widehat{y}}_K^-]$). We then predict the click probabilities for the entire set $S$ consisting of both positive and negative samples. 

To compute the prediction error, we adopt the pseudo-rank score as proposed in \cite{hl2}. The pseudo-rank score for each instance is defined as:

\begin{equation}
p_i = \frac{\exp(\widehat{y}_i^+)}{\exp(\widehat{y}_i^+) + \sum_{j=1}^K \exp(\widehat{y}_{i,j}^-)}
\label{eq17}
\end{equation}

The final NCE loss is computed as the cross-entropy loss over the pseudo-rank scores for all training instances in set $S$:

\begin{equation}
L = -\sum_{i \in S} \log(p_i)
\label{eq18}
\end{equation}

where $S$ is the set of the positive training samples.

\section{IV. MIND-tiny}
\label{sec:III.MIND-tiny}

\subsection{4.1 Overview of the MIND Dataset}

\begin{table}[!htbp]
\centering
\begin{tabular}{lcc}
\hline
\textbf{Statistic} & \textbf{MIND-small} & \textbf{MIND-large} \\
\hline
\#News & 65,238 & 161,013 \\
\#Categories & 18 & 20 \\
\#Impressions & 230,117 & 15,777,377 \\
\#Clicks & 347,727 & 24,155,470 \\
\hline
\end{tabular}
\caption{Statistics of the original MIND dataset. \textit{Table adapted from studies} \cite{rw8}.}
\label{tab:mindstats}
\end{table}

The Microsoft News Dataset (MIND) serves as a comprehensive benchmark for evaluating news recommendation systems \cite{rw8}. It contains millions of news articles and user click histories collected from the Microsoft News platform. The dataset is bifurcated into two primary subsets MIND-small and MIND-large. As delineated in Table \ref{tab:mindstats}, MIND-small contains over 65K news articles and approximately 230K user impressions, while MIND-large expands this to over 161K articles and nearly 16 million impressions. The dataset provides rich textual content including titles, abstracts, categories, and subcategories. MIND highly suitable for tasks in natural language processing and user modeling.

\subsection{4.2 Motivation and Construction of MIND-tiny}

While the full-scale MIND dataset is highly comprehensive, its size poses significant challenges for training on resource-constrained devices. To overcome this limitation while retaining the essential properties of large-scale user behavior and linguistic diversity, we introduce \textbf{MIND-tiny}—a compact yet representative dataset tailored for efficient model training and experimentation. The construction of MIND-tiny is guided by two key objectives: (1) to preserve temporal behavioral patterns of users, and (2) to retain representative and frequently interacted news content. To achieve this, we apply a selective sampling strategy from MIND-large based on the following criteria:

\begin{itemize}
\item \textbf{High-activity users:} We select users who have exhibited a minimum of 24 click interactions. Our hypothesis is that these high-activity users provide richer temporal behavior sequences, allowing models to better learn and generalize user interest dynamics over time.
\item \textbf{Popular news filtering:} Among the news articles, we retain those that were clicked by a large number of users (on average more than 150 times), sorted in descending order of popularity. This ensures that the dataset captures widely shared and semantically rich news content, representative of real-world reading trends.
\end{itemize}

By combining these filters, MIND-tiny preserves the essential behavioral and semantic characteristics of MIND-large while significantly reducing the overall data size. This strategy allows us to balance computational efficiency with the retention of critical information patterns from the original large-scale MIND dataset, making it well-suited for training model.\footnote{\textit{Filter high-activity users may underrepresent casual readers, potentially biasing results and limiting generalizability; future work should balance efficiency with broader audience diversity.}}

\subsection{4.3 Dataset Statistics and Analysis}
Detailed statistics of MIND-tiny are presented in Table \ref{tab:mindtiny-stats}. The dataset comprises 4,844 users and 19,557 unique news articles. It includes 16 top-level categories and 23 subcategories. On average, the abstract length is approximately 30 words, while titles are concise, with an average of 10.88 words.

\begin{table}[!htbp]
\centering
\begin{tabular}{ll}
\hline
\textbf{Statistic} & \textbf{Value} \\
\hline
\#News & 19,557 \\
\#Users & 4,844 \\
\#Categories & 16 \\
\#Subcategories & 23 \\
Avg. Title Length & 10.88 words \\
Avg. Abstract Length & 30.02 words \\
\hline
\end{tabular}
\caption{Statistics of the MIND-tiny dataset.}
\label{tab:mindtiny-stats}
\end{table}

\begin{figure}[ht]
\centering
\includegraphics[width=\columnwidth]{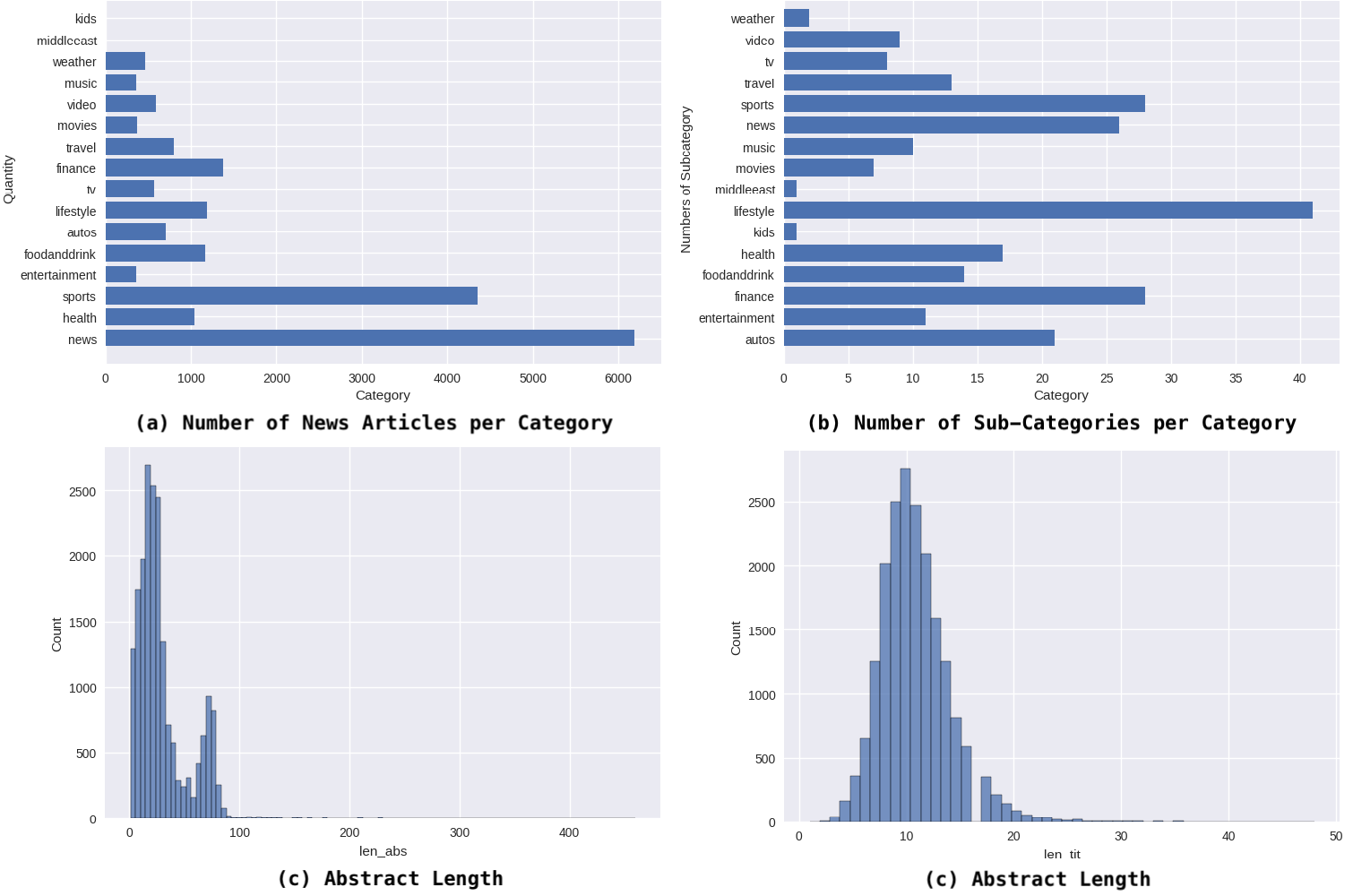}
\caption{Distributional statistics of categories, title lengths, and abstract lengths in MIND-tiny.}
\label{fig1}
\end{figure}

As illustrated in Figure \ref{fig1}, the dataset has a high proportion of news in the \textit{sports} and \textit{news} categories (approximately 4,300 and 6,000 articles respectively). The subcategory distribution is highly diverse, supporting varied semantic signals for content-based modeling. Abstract lengths show a bimodal distribution, with major peaks between 0–40 and 60–80 words, indicating a mix of short and longer summaries. Titles, in contrast, are mostly short and centered around 5–15 words. This distribution highlights the importance of combining content views (e.g., title and abstract) for improved understanding.

\section{V. Experiments}
\subsection{5.1 Experiment Setup}
\subsubsection{Dataset.} We conduct our experiments using the MIND dataset \cite{rw8}, a real-world news recommendation dataset collected from user behavior logs on Microsoft News. It is available in two versions: \textbf{MIND-large}, which contains over 15 million impression logs generated by 1 million users, and \textbf{MIND-small}, a subset randomly sampled from 50,000 users, as detailed in Table~\ref{tab:mindstats}. During training, we combine MIND-small and MIND-tiny to form the training set. Evaluations are conducted separately on MIND-small and MIND-large to assess generalizability across different data scales. See Section III.MIND-tiny for a summary of dataset statistics.

\begin{table*}[ht]
\centering
\caption{Performance of different methods on MIND-small and MIND-large. The best results are highlighted in bold, second is underlined (higher is better). §: Results from \cite{li2022miner}. ‡: MIND-large results from \cite{rw8}. †: Results from \cite{qi2021hierec}.}
\resizebox{\textwidth}{!}{
\begin{tabular}{l|c|cccc|cccc}
\hline
& & \multicolumn{4}{c|}{\textbf{MIND-small}} & \multicolumn{4}{c}{\textbf{MIND-large}} \\
\textbf{Model} & \textbf{\#Params} & AUC & MRR & nDCG@5 & nDCG@10 & AUC & MRR & nDCG@5 & nDCG@10 \\
\hline
LibFM§ & - & 0.5974 & 0.2633 & 0.2795 & 0.3429 & 0.6185 & 0.2945 & 0.3145 & 0.3713 \\
DeepFM§ & - & 0.5989 & 0.2621 & 0.2774 & 0.3406 & 0.6187 & 0.2930 & 0.3135 & 0.3705 \\
\hline
NRMS‡ & 22M & 0.6183 & 0.2753 & 0.2980 & 0.3653 & 0.6776 & 0.3305 & 0.3594 & 0.4163 \\
Hi-Fi Ark & 21.9M & 0.6049 & 0.2647 & 0.2927 & 0.3560 & - & - & - & - \\
NPA§ & - & 0.6465 & 0.3001 & 0.3314 & 0.3947 & 0.6592 & 0.3207 & 0.3472 & 0.4037 \\
TANR & 21.7M & 0.6338 & 0.2868 & 0.3169 & 0.3804 & - & - & - & - \\
LSTUR‡ & 71.6M & 0.6372 & 0.2769 & 0.2559 & 0.3099 & 0.6773 & 0.3277 & 0.3559 & 0.4134 \\
NAML‡ & 22.2M & 0.6195 & 0.2528 & 0.2675 & 0.3356 & 0.6686 & 0.3249 & 0.3524 & 0.4091 \\
DKN‡ & 22.9M & 0.5954 & 0.2659 & 0.2909 & 0.3518 & 0.6460 & 0.3132 & 0.2909 & 0.3518 \\
HieRec† & - & \underline{0.6795} & \underline{0.3287} & \underline{0.3636} & \underline{0.4253} & 0.6903 & 0.3389 & 0.3384 & 0.3948 \\
MINER§ & - & \textbf{0.6961} & \textbf{0.3397} & \textbf{0.3762} & \textbf{0.4390} & \textbf{0.7151} & \textbf{0.3618} & \textbf{0.3972} & \textbf{0.4534} \\
\hline
Co-NAML-LSTUR & 46.4M & 0.6571 & 0.3119 & 0.3465 & 0.4028 & \underline{0.6931} & \underline{0.3420} & \underline{0.3698} & \underline{0.4275} \\
\hline
\end{tabular}
}
\label{tab:main-results}
\end{table*}

\subsubsection{Training Configuration.} Following prior work \cite{wu2019neural}, we utilize the most recent 60 news clicked by a user to construct user representations. We use only the news title, abstracts (up to 64 tokens) for most experiments. For models that support abstract and category features (e.g., NAML, LSTUR), we additionally use news categories, and subcategories (maximum length of 16 tokens).
All models are trained end-to-end, with the exception of the DistilBERT encoder, which is frozen during training and used solely for extracting word-level semantic embeddings. We adopt a negative sampling ratio of $K=3$, meaning each positive sample is paired with 3 negative news candidates. We use a batch size of 128 and train for 5 epochs with a learning rate of $1 \times 10^{-4}$, using the Adam optimizer \cite{kingma2014adam}. The dropout probability is set to 0.3 to prevent overfitting. All experiments are conducted on a single NVIDIA RTX A6000 GPU over a training period of 36 hours.

Due to hardware constraints, we conduct Section 5.4 (\textit{Ablation Study}) and Section 5.5 (\textit{Effectiveness of Click Prediction}) evaluations (e.g., Table~\ref{tab:ablation}, Table~\ref{tab:click-predictor}) using the validation splits of the MIND-small dataset. We follow standard practice \cite{rw8} and adopt four widely-used ranking metrics for performance evaluation: AUC, MRR, nDCG@5, and nDCG@10. These metrics capture the ranking quality of the recommended news articles for each user impression.

\subsection{5.2 Comparison Methods}
We compare our approach against two categories of baseline models: 

\textbf{Feature-based Methods:} Traditional models relying on manual feature engineering and collaborative filtering, including:  
\textbf{(1) LibFM} \cite{rw4}: A factorization machine using user ID, news ID, and TF-IDF of news titles. This model is effective for sparse data but relies heavily on handcrafted features.  
\textbf{(2) DeepFM} \cite{rw5}: A hybrid of factorization machines and deep neural networks applied to the same features. It captures both low- and high-order feature interactions without manual design.  

\textbf{Neural Recommendation Methods:} Deep learning-based architectures tailored for news recommendation, including  
\textbf{(3) DKN} \cite{bib3}: A knowledge-aware model combining CNN with entity embeddings to enrich the semantic representation of news.  
\textbf{(4) NPA} \cite{wu2019npa}: A personalized attention-based model that adaptively selects informative words and news based on individual user interests.  
\textbf{(5) NAML} \cite{bib1}: A multi-view learning framework that jointly models title, abstract, and category information with attentive pooling.  
\textbf{(6) LSTUR} \cite{an2019neural}: A model that distinguishes short- and long-term user interests using GRUs and user embeddings.  
\textbf{(7) NRMS} \cite{rw13}: A self-attention-based model that captures sequential user preferences and contextual dependencies between news.  
\textbf{(8) HiFiArk} \cite{liu2019hi}: A hierarchical architecture that models fine-grained user preferences across subtopic and topic levels.  
\textbf{(9) TANR} \cite{wu2019neural}: A model that jointly learns topic-aware news and user representations using latent topics inferred from user behavior.  
\textbf{(10) HieRec} \cite{qi2021hierec}: A hierarchical user modeling framework that encodes preferences at multiple levels for better personalization.  
\textbf{(11) MINER} \cite{li2022miner}: A modular architecture that integrates multiple views and selectors to dynamically capture diverse user reading patterns.  

\subsection{5.3 Main Results}

Table~\ref{tab:main-results} presents the performance of Co-NAML-LSTUR against a broad range of baseline methods on both MIND-small and MIND-large datasets. We compare against both traditional recommendation methods (LibFM, DeepFM) and a range of neural news recommenders, including content-based models (NRMS, LSTUR, NAML, DKN), attention-based models (NPA, TANR), and recent state-of-the-art frameworks (HieRec, MINER). All metrics are presented in percentage form (without \%) for readability, and the best and second-best results are bolded and underlined, respectively.

On the MIND-small dataset, Co-NAML-LSTUR achieves an AUC of 0.6571 and an MRR of 0.3119, surpassing several strong baselines such as NRMS (+4.2\% AUC), LSTUR (+2.0\% MRR), and NAML (+4.0\% nDCG@10), highlighting its effective modeling of both semantic and temporal user signals. While MINER yields the best overall performance, Co-NAML-LSTUR attains the second-highest scores across all metrics.

On MIND-large, Co-NAML-LSTUR maintains robust performance and generalizes well with an AUC of 0.6931 and nDCG@10 of 0.4275, outperforming nearly all neural baselines. It significantly improves over classical baselines and remains on par with MINER across all metrics. Our model shows consistent improvements over NRMS (+1.55 AUC, +1.15 MRR) and NAML (+2.45 AUC, +1.71 MRR). These results confirm the robustness of our dual-view encoder and the effectiveness of integrating both short- and long-term user interests under a unified attention framework.

\paragraph{Complementarity of design choices.} Our improvements can be attributed to two key design principles: (1) multi-view news encoding that captures diverse content signals (e.g., title, abstract, category), and (2) hierarchical user modeling that dynamically balances recent clicks and long-term interests. Unlike LSTUR, which uses a GRU-based user encoder with limited context fusion, Co-NAML-LSTUR enables richer user representation via cross-view attention. This leads to better generalization, particularly in settings with longer histories or sparse clicks.

\noindent\textbf{Comparison with state-of-the-art models.} Although MINER reports the highest performance on both benchmarks, it requires modeling multiple user interest vectors and computing adaptive selection weights, which leads to greater processing overhead. In contrast, Co-NAML-LSTUR achieves a competitive balance between accuracy and processing efficiency; our model—although trained on fewer resources by concatenating data from MIND-small and MIND-tiny—still delivers performance comparable to MINER, which is trained on the much larger MIND-large dataset. Overall, these results validate the core hypothesis of this work: that incorporating multi-view news semantics with an attentive long- and short-term user modeling pipeline yields meaningful performance benefits.

Overall, these results validate the core hypothesis of this work: that incorporating multi-view news semantics (title, abstract, category) with an attentive long- and short-term user modeling pipeline yields meaningful performance benefits. Co-NAML-LSTUR provides a strong balance of accuracy, parameter efficiency (46.4M), and extensibility, making it a competitive solution for real-world news recommendation tasks.

\subsection{5.4 Ablation Study}
To better understand the contributions of individual components within our Co-NAML-LSTUR architecture, we perform an ablation study on the MIND-small dataset. Table~\ref{tab:ablation} presents the performance impact of removing or altering key modules, with evaluation metrics including AUC, MRR, and nDCG@10. In Co-NAML-LSTUR w/o BERT, we use word embedding through the pre-trained Glove \cite{pennington2014glove} vectors.

\begin{table}[!htbp]
\centering
\resizebox{\columnwidth}{!}{
\begin{tabular}{lccc}
\hline
\textbf{Model} & \textbf{AUC} & \textbf{MRR} & \textbf{nDCG@10} \\
\hline
\textbf{Co-NAML-LSTUR (Full)} & \textbf{0.6571} & \textbf{0.3119} & \textbf{0.4028} \\
\quad w/o Category Embedding & 0.6400 & 0.2980 & 0.3870 \\
\quad w/o Word Embedding & 0.6105 & 0.2760 & 0.3600 \\
\hline
\textbf{Co-NAML-LSTUR w/o BERT} & \textbf{0.6002} & \textbf{0.2529} & \textbf{0.3429} \\
\quad w/o Category Embedding & 0.5875 & 0.2450 & 0.3305 \\
\quad w/o Word Embedding & 0.5650 & 0.2320 & 0.3100 \\
\hline
\textbf{NAML + BERT (Baseline)} & \textbf{0.6350} & \textbf{0.2850} & \textbf{0.3750} \\
\hline
\end{tabular}}
\caption{Ablation results on MIND-small. Removing individual components consistently degrades performance across all metrics.}
\label{tab:ablation}
\end{table}

We observe that the full Co-NAML-LSTUR model achieves the highest performance across all metrics. Removing the BERT-based word embeddings leads to a substantial drop in AUC (from 0.6571 to 0.6002) and nDCG@10 (from 0.4028 to 0.3429), demonstrating the effectiveness of leveraging contextualized embeddings for the semantic representation of news content. This confirms that BERT captures deeper linguistic patterns compared to static word embeddings.

Eliminating category and subcategory embeddings also results in notable performance degradation. The decrease in AUC (by 1.7–2.5 points depending on whether BERT is used) suggests that these structured metadata features complement textual views by providing high-level semantic signals. This highlights the importance of integrating both unstructured and structured news information.

Interestingly, the impact of removing word-level embeddings is even more pronounced in the non-BERT setting (drop from 0.6002 to 0.5650 in AUC), suggesting that without BERT, the model becomes more reliant on traditional word-level semantics. This confirms the complementary role of BERT and conventional embeddings when jointly used in a multi-view encoding pipeline. The ablation results validate the design of Co-NAML-LSTUR, in which each component—BERT-enhanced word embeddings, category-level metadata, and multi-view fusion—contributes to performance improvements.

\subsection{5.5 Effectiveness of Click Prediction}
As described in Section 3.3 (\textit{Click Predictor}), our framework supports two variants of the click prediction module: a simple dot-product function and a neural network-based predictor. We evaluate both variants under the NAML (as Baseline) and Co-NAML-LSTUR settings to analyze their relative effectiveness and computational trade-offs. The results on MIND-small are summarized in Table~\ref{tab:click-predictor}.

\begin{table}[!htbp]
\centering
\resizebox{\columnwidth}{!}{
\begin{tabular}{lccc}
\hline
\textbf{Model} & \textbf{AUC} & \textbf{MRR} & \textbf{nDCG@10} \\
\hline
NAML w/o Neural Network & 0.6195 & 0.2528 & 0.3356 \\
NAML with Neural Network & 0.6402 & 0.2858 & 0.3819 \\
\hline
Co-NAML-LSTUR w/o Neural Network & 0.5986 & 0.2513 & 0.3406 \\
Co-NAML-LSTUR with Neural Network & \textbf{0.6571} & \textbf{0.3119} & \textbf{0.4028} \\
\hline
\end{tabular}
}
\caption{Effect of click prediction module design on recommendation performance. The neural predictor consistently outperforms the dot-product across all evaluation metrics.}
\label{tab:click-predictor}
\end{table}


Across both model backbones, the neural network-based click predictor significantly outperforms the dot-product variant in AUC, MRR, and nDCG@10. Specifically, under the NAML backbone, integrating a neural predictor improves nDCG@10 from 0.3356 to 0.3819, demonstrating enhanced modeling of complex user-item interactions. The improvement is even more pronounced in the Co-NAML-LSTUR configuration, with AUC increasing by over 5.8 points (from 0.5986 to 0.6571) and nDCG@10 reaching 0.4028.

This performance gain can be attributed to the higher capacity of the neural click predictor, which can capture non-linear relationships between user and news representations beyond what dot-product similarity can express. However, we note that this improvement comes at a computational cost: the neural predictor requires approximately 2X more training time to converge compared to the dot-product variant. This highlights a trade-off between performance and efficiency, depending on deployment requirements. Our results confirm that a learnable neural click predictor significantly enhances click prediction quality and overall recommendation performance, particularly when paired with a strong encoder such as Co-NAML-LSTUR.

\subsection{5.6 Case Study and Visualization}

To provide a deeper understanding of our model’s recommendation behavior, we conduct a qualitative case study and visualization analysis based on user interactions in the test set. Specifically, we compare the predicted results of \textit{Co-NAML-LSTUR} with the user’s actual click history to examine how well the model captures diverse user interests.

\begin{figure}[ht]
\centering
\includegraphics[width=\columnwidth]{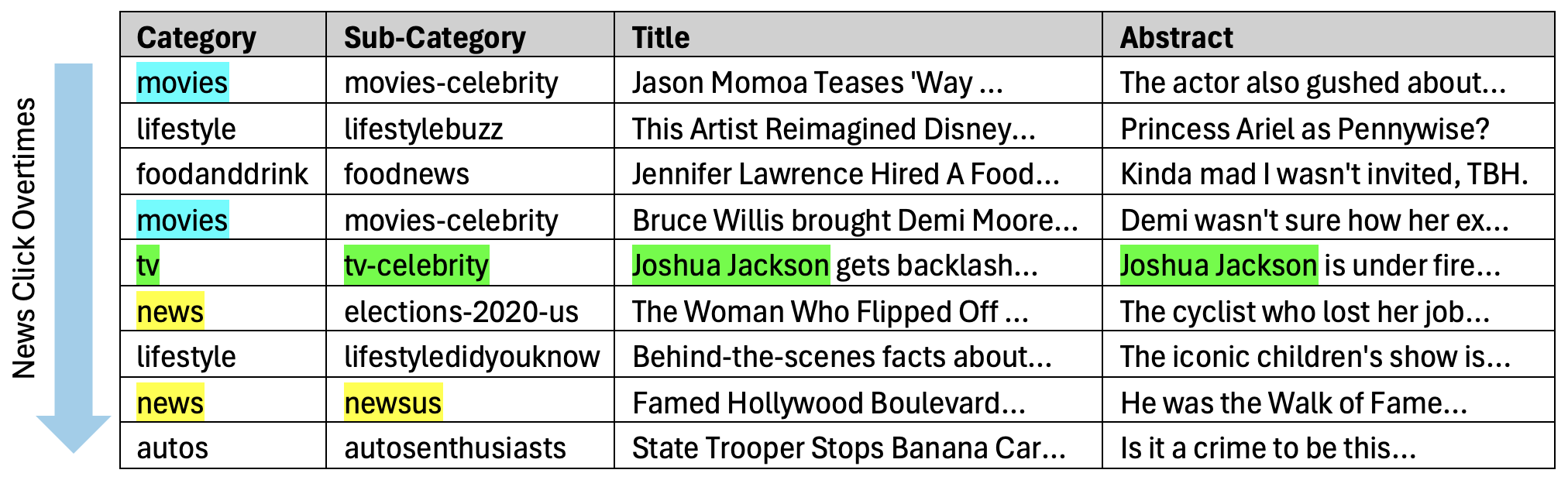}
\caption{The news click history of a sampled user, who has
various interests including entertainment, lifestyle and food.}
\label{fig:user_history}
\end{figure}

Figure~\ref{fig:user_history} presents the historical click behavior of a randomly selected user, where we visualize the nine most recent news articles the user clicked. The user's history spans a range of topics, including entertainment (e.g., \textit{movies}, \textit{tv-celebrity}), lifestyle, and food. This illustrates the user’s multi-faceted interests rather than a single dominant preference.

\begin{figure}[ht]
\centering
\includegraphics[width=\columnwidth]{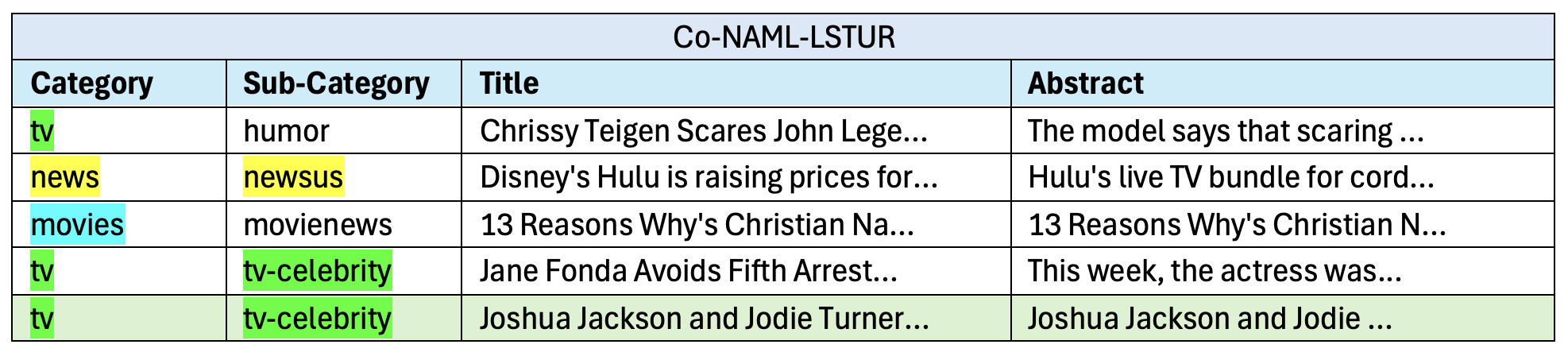}
\caption{Top-5 news recommendations from \textit{Co-NAML-LSTUR}. The green-highlighted row indicates an article that the user actually clicked.}
\label{fig:model_prediction}
\end{figure}

Figure~\ref{fig:model_prediction} shows the top-5 news recommendations generated by the \textit{Co-NAML-LSTUR} model for this user. We observe that the model accurately prioritizes news from categories previously clicked by the user, such as \textit{tv-celebrity} and \textit{movies}. For example, a predicted article about the actress Jane Fonda aligns closely with the user’s prior interest in celebrity news. Furthermore, the model also surfaces content from the \textit{news} category, which, although less frequent in the user’s history, still appears to be contextually relevant based on the user’s broader behavior.

These results highlight the model’s ability to effectively model nuanced user interests and diversify recommendations beyond the most frequently clicked category. Additionally, the incorporation of multi-head user representations and content-aware attention mechanisms allows the model to balance between topical relevance and content novelty. Notably, the user clicked on one of the recommended articles (highlighted in green), providing further evidence of the model’s practical effectiveness.

This case study qualitatively confirms that our model does not simply overfit to dominant categories but is capable of generalizing across related interest areas, leveraging both user behavior patterns and content semantics.

\section{VI. Conclusion}

In this paper, we propose Co-NAML-LSTUR, a unified and modular neural framework for personalized news recommendation. Our approach addresses key challenges in modeling both user interest diversity and the semantic richness of news content. To represent news articles, we design a hybrid encoder that integrates category-aware and content-sensitive features through attentive multi-view learning. On the user side, we fuse long- and short-term user preferences by leveraging the strengths of both static profile embeddings and sequential browsing behavior modeling, allowing the model to adapt to dynamic reading patterns. A central component of our framework is the incorporation of a click prediction module, which can flexibly switch between dot-product and neural interaction functions. Our experiments demonstrate that the neural variant significantly enhances recommendation accuracy, albeit with increased computational cost. We conduct extensive evaluations on the MIND dataset, showing that Co-NAML-LSTUR outperforms several state-of-the-art baselines across multiple ranking metrics. Additionally, our case study and visual analysis reveal that the model is capable of capturing multi-faceted user interests and generating relevant, diverse recommendations. Our method makes it a promising foundation for future work. It can be extended with external knowledge, multi-modal content, or user demographic features to further boost performance and interpretability (such as analyze attention weight distributions, feature importance, and user group variations, provides quantitative interpretability or user-behavior analysis).

\subsubsection{Authorship contribution statement.} Minh Hoang Nguyen: Methodology, Conceptualization, Writing – structuring original draft, Writing—reviewing. Thuat Thien Nguyen: Preprocessing data, Building dataset, Conceptualization, Writing – structuring original draft. Minh Nhat Ta: Preprocessing data, Writing – structuring original draft. Tung Le: Supervision, Supporting, Reviewing and Editing. Huy Tien Nguyen: Supervision, Supporting, Conceptualization, Project Administration.

\subsubsection{Acknowledgements.} This research is partially funded by the Vingroup Innovation Foundation (VINIF) under the grant number VINIF.2021.JM01.N2. 

This preprint has not undergone peer review (when applicable) or any post-submission improvements or corrections. The Version of Record of this contribution is published in Multi-disciplinary Trends in Artificial Intelligence (MIWAI 2025), Lecture Notes in Computer Science, vol. 16354, and is available online at \underline{\url{https://doi.org/10.1007/978-981-95-4960-3_9}}.

\subsubsection{Disclosure of Interests.} The authors declare that they have no known competing financial interests or personal relationships that could have appeared to influence the work reported in this paper.

\bigskip
\bibliography{refs}

\end{document}